\documentclass[conference]{IEEEtran}
\usepackage{cite}
\usepackage{amsmath,amssymb,amsfonts}
\usepackage{algorithmic}
\usepackage{graphicx}
\usepackage{textcomp}
\usepackage{xcolor}
\usepackage{booktabs}
 
\def\BibTeX{{\rm B\kern-.05em{\sc i\kern-.025em b}\kern-.08em
    T\kern-.1667em\lower.7ex\hbox{E}\kern-.125emX}}
\begin{document}

\title{Potential Ways to Detect Unfairness in HRI and to Re-establish Positive Group Dynamics}

\thanks{anonymized}

\author{\IEEEauthorblockN{Astrid Rosenthal-von der Pütten}
\IEEEauthorblockA{\textit{iTec - Chair Individual and Technology} \\
\textit{RWTH Aachen University}\\
Aachen, Germany \\
arvdp@itec.rwth-aachen.de}
\and
\IEEEauthorblockN{Stefan Schiffer}
\IEEEauthorblockA{\textit{iTec - Chair Individual and Technology} \\
\textit{RWTH Aachen University}\\
Aachen, Germany \\
stefan.schiffer@itec.rwth-aachen.de}
}

\maketitle

\begin{abstract}
This paper focuses on the identification of different algorithm-based biases in robotic behaviour and their consequences in human-robot mixed groups. We propose to develop computational models to detect episodes of microaggression, discrimination, and social exclusion informed by a) observing human coping behaviours that are used to regain social inclusion and b) using system inherent information that reveal unequal treatment of human interactants. Based on this information we can start to develop regulatory mechanisms to promote fairness and social inclusion in HRI. 
\end{abstract}


\section{Introduction}
Social robots are envisioned to be part of our lives as service providers, team members, companions. Depending on the robots' tasks and purpose they will be playing a more or less active role in our social groups and potentially shape group dynamics - for the better and the worse. Previous research demonstrated that robots can positively influence social dynamics in small groups. In free play situations a robot was able to mitigate conflict between children over toys by providing information on how to compromise \cite{shen2018stop}. A robotic microphone positively influenced group discussions by encouraging those discussion partners to participate more who were more silent than others \cite{tennent2019micbot}. Similarly, a robot giving self-disclosure statements could facilitate to speak up in a support group session for stressed students and it improved perceptions of trust among the members of the support group \cite{birmingham2020can}. However, robots can also cause feelings of social exclusion by leaving humans out of interactions (e.g. not tossing a ball to the human, \cite{erel2021excluded}) or communication (e.g. speaking in "robotic language, \cite{rosenthalvonderpuetten2023socialex}, or bluntly rejecting the human, \cite{nash2018bionic}), causing negative consequences such as experiencing negative emotions, the feeling of being ignored or being meaningless, and lowered self-esteem. While we assume that developer teams of social robots do not intend to create robots that socially exclude individuals, social exclusion can still arise in interactions in human-robot groups because robots may have software components that are biased against certain groups of humans (e.g., women, PoC) or because it is unaware of the social situation the robot finds itself in and unknowingly behaves socially inadequate. 

In this paper we want to briefly revisit i) the role groups play in our (human) lives and how group membership can lead to inter-group bias, ii) the psychological consequences of social rejection (caused by biased behavior), iii) sources of algorithmic bias, and iv) how to use system information to detect bias and start repair mechanisms.

\section{Related work on HRI groups and algorithmic bias}
\subsection{What groups mean to us}
Groups are highly important to individuals \cite{baumeister1995need}. Since the membership in groups is one defining part of an individual’s self-concept and consequently an individual’s self-esteem is partly dependent upon group membership, strategies to protect the group and differentiate it from other groups are important for the individual. Positive distinctiveness of the in-group from other groups can be achieved by simply evaluating groups differently in favour of the in-group – also referred to as inter-group bias which is “the systematic tendency to evaluate one’s own membership group (the in-group) or its members more favourably than a non-membership group (the out-group) or its members” \cite{hewstone2002intergroup}. It manifests as favouring the in-group (in-group favouritism) or derogating the out-group (out-group derogation), or both. In-group favouritism entails the extension of trust, positive regard, cooperation, and empathy to in-group members, but not to members of the out-group and thus is an initial form of discrimination. Inter-group bias extends to robots. For instance, humans show in-group favouritism for an in-group robot in online studies \cite{eyssel2012if,kuchenbrandt2011minimal} and assigned “painful” noise blasts to out-group humans to spare in-group robots in scenarios were interactants were in different rooms \cite{fraune2020some}. Since humans show inter-group bias in human-robot mixed groups negative emotional and social consequences potentially arise for other humans when a robot is favoured instead of them. Moreover, the robots could also be the source of social rejection due to algorithmic biases which will be discussed further below.

\subsection{What happens when we feel excluded from a group}
Inter-group bias can be perceived as a sign of social exclusion or social rejection. According to the Temporal-Need-Threat-Model by Williams \cite{williams2009ostracism}, social exclusion causes a reflexive pain response accompanied with negative affect (e.g., sadness, anger) and triggers threats to four fundamental needs: belonging, self-esteem, control over one’s social environment and meaningful existence. In a reflective stage, individuals’ attention is directed to the exclusion episode and they reflect on its meaning and relevance. This may lead to coping responses such as compliance and conformity or attracting attention, provoking, and attempts of controlling others to fortify the threatened needs. Persistent exposure to ostracism over time consumes the resources necessary to motivate the individual to fortify threatened needs. Eventually, this leads to resignation, alienation, helplessness, and depression. Since humans are hypersensitive to ostracism and tend to over-detect ostracism \cite{zadro2004low}, it is extremely likely that humans detect ostracism in interactions with robots as well and experience and engage in the described reflexive and reflective processes. Indeed, recent studies have explored this and found that participants felt excluded when robots talked in a “robot language” \cite{rosenthalvonderpuetten2023socialex} or when a robot stated it did not want to interact with the human again \cite{nash2018bionic}. Although the need for a paradigm shift from studying dyadic human-robot interactions in laboratory settings to studying group interactions in complex environments has been identified and advocated for \cite{jung2018robots} research in human-robot mixed groups is still scarce. Social psychological phenomena such as social exclusion, and ostracism as negative consequences of a robot’s unequal adaptation to group members through machine learning is yet a new perspective in research on HRI groups that the community just recently has identified to be important.

\subsection{Sources of algorithmic bias in HRI}
The general notion of unfair or fair AI has been discussed intensively in recent years. In our modern, digitalized world, we engage more and more in interactions with algorithms and artificially intelligent systems that learn and adapt based on these interactions. Our visits, views, clicks, and buying decisions provide training data for recommender systems on shopping websites (e.g., Amazon) or video streaming applications (e.g., Netflix). Recently, voice agents have entered our homes providing us with helpful information, and services while using these interactions as training data to learn and adapt to us and generalizing this knowledge to predict preferences and intentions of groups of users. Especially in the latter area of voice agents, similar biases may emerge when algorithms try to categorize users into groups and provide these groups with personalized interactions. Recent research demonstrated in many application fields (e.g., financial credit, job application management) that algorithms often discriminate certain groups of people, for instance, based on gender or skin tone and thereby exhibit unintended and unexpected biases usually originated in biased training data. While the lack of diversity in the training data sets that are being used in machine learning originates from different sources, it unequivocally causes a bias towards certain types of users at the cost of others. A new topic that has been recently identified \cite{howard2018ugly} are potential negative consequences arising in HRI by robots that show unintended biases in favour of certain group members and thereby discriminating others. Under the term Fair AI, researchers call out the computer science community to “identify sources of bias, [to] de-bias training data and [to] develop artificial-intelligence algorithms that are robust to skews in the data” \cite{zou2018ai}. Since computer vision and machine-learning are core technologies for robotic systems, it has been proposed that a similar threat is posed to HRI \cite{righetti2019unintended}. 

Interestingly, concerns about the negative effects of biased robotics systems are often seen from a more global societal perspective. For instance, autonomous cars could put people of colour to a greater risk due to biased person recognition and medical or service robots might reinforce certain discriminatory practices due to biased algorithmic decision-making \cite{howard2018ugly}. However, besides the issues already identified, new forms of biases are likely to emerge when the training data base for machine learning are interactions with multiple humans over a longer time as we have discussed in previous work \cite{rosenthal2020social}. Robots are expected to learn and adapt to their users, ideally while in operation during run-time. Hence, robots learning from humans means that robots learn from interactions and the more interactions the better the learning outcome. But humans might have more or less time or might be more or less motivated to provide these interactions that are needed for learning. Thus, training data sets differ in quantity and quality which has consequences for the learning outcome (e.g., knowing the user’s preferences) and the robot’s quality to adapt to different users. 

Let is consider the following family scenario, in which the user who spends more time at home potentially provides the largest training data base for the robot, is best known to the system, and his/her preferences can be easily determined and served. A user who spends less time at home might receive recommendations and interactions matching his/her preferences less often. Or let us consider a working environment, in which the robot’s implemented goal is to maximize team performance. The robot will monitor the performance of every single team member and their contribution to team performance. Based on the maximization goal, the robot might decide to distribute more resources to those team members who are high performers in the task, thereby discriminating low performers. Very likely, low performers will experience negative emotions, feel threatened in their self-esteem and their need to belong to the group, and will try to regain social inclusion. Recent work tapped into this issue of unequal adaptation to users based on performance and algorithm goals in experimental studies. For instance, in a collaborative tower construction task, a robot distributed building blocks unequally between two participants which led to lower satisfaction of the human team members with the team relationship \cite{jung2020robot}. In a collaborative Tetris Game, fair distribution (in contrast to unfair distribution) of resources led participants to trust the system more and resulted in higher overall team performance \cite{claure2020multi}. However, emotional responses and consequences for the self-perception and self-esteem of the neglected participant were not assessed.

These first results and the scenarios described above demonstrate that besides the now commonly known problems of biases in natural language processing or face recognition also interaction-based algorithmic learning can result in, for instance, perceived (inter-group) bias and social exclusion of individuals with severe negative outcomes for the emotional state of the individual and the social dynamics of the group. 

\section{How to overcome biased HRI and reach better inclusion}

\subsection{First Step - Recognizing the potential for biases in your own work}
Researchers in the field of HRI have become more aware of the potential that their developments and systems might be affected by biases. Earlier this year a group of HRI scholars discussed "how pursuing a very typical, data-driven approach to the development of a robot listener behavior (production of backchannels, which can serve to indicate attentiveness) resulted in models that acted differently with participants with different gender identities" \cite{parreira2023did}. In their paper the authors discuss design guidelines that may be applied to avoid embedding gender biases into robot social behavior such as carefully examining training data sets before using them for modelling. According to Ntoutsi et al. \cite{ntoutsi2020bias} this recommendation would fall under preprocessing methods focusing on the data to mitigate bias which focus on creating so-called balanced data sets. This can be done using different approaches such as equal sampling from different groups or altering the given data in its classification, i.e., adapting training sample weights \cite{krasanakis2018adaptive}. 

\subsection{Second Step - Mitigating bias in machine learning before system deployment}
Besides the pre-processing methods to mitigate bias as mentioned before, Ntoutsi et al. \cite{ntoutsi2020bias} also consider so-called in-processing methods focusing on the ML algorithm, and post-processing methods focusing on the ML model. Both types of approaches concentrate on the machine learning process and/or the inspection and adaptation of the resulting model. For instance, in the latter case Ntoutsi et al. refer to previous work that post-hoc changed the confidence of CPAR classification rules \cite{pedreschi2009measuring} or the probabilities in Naïve Bayes models \cite{calders2010three}. 

\subsection{Third Step - How to use system information to detect bias during interaction and start repair mechanisms }
All the specified approaches above have in common that developers or researchers are actively involved in curating either data or changing the algorithm's specifications which cannot be done during run-time. Moreover, if the system is further learning based on continuous interactions it can be that a "de-biased" algorithm becomes biased again, for instance, because human interactants behave in stereotypical ways. 
We have proposed that i) information on biased components and ii) certain system information that is produced during ongoing interactions with humans can be used to inform the system about potential biases emerging. For instance, it is commonly known that biased speech recognition performance is biased in favour for people speaking accent-free standard languages due to better training to that user type. This known pre-existing bias can be used during the development of interactions with humans. The system should also be enabled to draw conclusions of internal data to detect bias. For instance, recognition for human faces or behaviours as well as predictions about human behaviour usually are hypothesis-based with specifications about the likelihood and confidence that this hypothesis is true or false. Consistent lower likelihoods connected with one user could be used (together with other information) as an indicator for bias. As described above most systems are biased in their speech recognition performance in favour of people speaking accent-free High German due to better training to that user type in contrast to people speaking local dialects or foreign accents. A robot could use system information that is an indicator for this bias occurring, for instance, when a higher number of hypotheses exists (cf. n-best lists, \cite{baumann2009evaluating}; \cite{verhasselt2010n}) and/or lower confidence (cf. \cite{breitenstein2009robust}) in speech recognition or computer vision for a specific user (e.g., interactant with local dialect or foreign language accent). Based on this the robot would initiate a regulatory mechanism such as apologizing for misunderstandings and asking the user to speak more slowly. 

\subsection{Fourth Step - How to use user behavior to detect bias during interaction and start repair mechanisms}
As explained above there is empirical evidence that when humans detect signs of social rejection or social exclusion by a robot they will experience negative emotions and fundamental needs are threatened \cite{nash2018bionic,erel2021excluded,erel2022carryover,rosenthalvonderpuetten2023socialex}. This may lead to coping responses such as compliance and conformity, attracting attention, provoking, or attempts of controlling others to fortify the threatened needs.
Current studies on social exclusion in HRI scenarios predominantly look into self-reported experiences. Future work should systematically investigate which coping behaviours excluded humans exert in trying to regain social inclusion. One approach is to use behaviour analysis to identify patterns of verbal and nonverbal behaviour, and interactional strategies a robot might detect as a sign that social exclusion occurred. 
Based on this classification a computational model could be implemented to detect episodes of social exclusion informed by observing human coping behaviours that are used to regain social inclusion as well as the aforementioned system information. 

\subsection{Fifth Step - Develop Socially Interactive Agents with Capacity to Re-establish positive Group Dynamics}
The work is not done when we managed to detect biases. We further need a good concept how to resolve social exclusion episodes for the human and re-establish positive group dynamics. This means that there is a need to i) develop conversational or interactional strategies to maintain positive social group dynamics that can be triggered when potential bias is detected, and ii) research which conversational and interactional strategies are effective and regarded as socially adequate in different situations and group constellations. 

\section{Conclusion}
In this paper we outlined why social robots should take into account the social dynamics in a human-robot mixed group as well as the (negative) social consequences of its own behaviour in these groups. We discussed why and in which ways biases can arise in HRI and how we can either de-bias systems or enable the system to automatically detect bias and engage in repair mechanisms. We advocate for considering this perspective throughout the development process of a new system.

\bibliographystyle{IEEEtran}
\bibliography{IEEEabrv,sample-base.bib}

\end{document}